\documentclass[letterpaper, 10 pt, conference]{ieeeconf}  % Comment this line out if you need a4paper

\IEEEoverridecommandlockouts                              % This command is only needed if 
\usepackage{hyperref}
\usepackage{url}
\usepackage{graphicx}
\usepackage{multirow} % added
\usepackage{listings} % added
\usepackage{amsmath}
\usepackage{setspace}

\title{\LARGE \bf
Learning Long Short-Term Intention within Human Daily Behaviors
}

\author{\hspace{0.2cm} Zhe Sun$^{1}$, Rujie Wu$^{2}$, Xiaodong Yang$^{3}$,  Hongzhao Xie$^{1}$, Haiyan Jiang$^{4}$, Junda Bi$^{5}$, Zhenliang Zhang$^{1*}$
\thanks{$^{*}$ Corresponding author.}
\thanks{$^{1}$ State Key Laboratory of General Artificial Intelligence, Beijing Institute for General Artificial Intelligence (BIGAI), China. {\tt\small \{sunzhe, xiehongzhao, zlzhang\}@bigai.ai}}
\thanks{$^{2}$ Peking University, China. {\tt\small wu\_rujie@stu.pku.edu.cn}}
\thanks{$^{3}$ University of Electronic Science and Technology of China, China. {\tt\small 202121081008@std.uestc.edu.cn}}
\thanks{$^{4}$ Beijing Institute of Technology, China. {\tt\small hy\_jiang@bit.edu.cn}}
\thanks{$^{5}$ Tsinghua University, China. {\tt\small bjd15@mails.tsinghua.edu.cn}}
}

\begin{document}

\maketitle
\thispagestyle{empty}
\pagestyle{empty}

%%%%%%%%%%%%%%%%%%%%%%%%%%%%%%%%%%%%%%%%%%%%%%%%%%%%%%%%%%%%%%%%%%%%%%%%%%%%%%%%
\begin{abstract}
In the domain of autonomous household robots, it is of utmost importance for robots to understand human behaviors and provide appropriate services. This requires the robots to possess the capability to analyze complex human behaviors and predict the true intentions of humans. Traditionally, humans are perceived as flawless, with their decisions acting as the standards that robots should strive to align with. However, this raises a pertinent question: What if humans make mistakes? In this research, we present a unique task, termed ``long short-term intention prediction''. This task requires robots can predict the long-term intention of humans, which aligns with human values, and the short term intention of humans, which reflects the immediate action intention. Meanwhile, the robots need to detect the potential non-consistency between the short-term and long-term intentions, and provide necessary warnings and suggestions. To facilitate this task, we propose a long short-term intention model to represent the complex intention states, and build a dataset to train this intention model. Then we propose a two-stage method to integrate the intention model for robots: i) predicting human intentions of both value-based long-term intentions and action-based short-term intentions; and 2) analyzing the consistency between the long-term and short-term intentions. Experimental results indicate that the proposed long short-term intention model can assist robots in comprehending human behavioral patterns over both long-term and short-term durations, which helps determine the consistency between long-term and short-term intentions of humans.

\end{abstract}

%%%%%%%%%%%%%%%%%%%%%%%%%%%%%%%%%%%%%%%%%%%%%%%%%%%%%%%%%%%%%%%%%%%%%%%%%%%%%%%%
\section{INTRODUCTION}

The symbiotic relationship~\cite{sandini2018social} between humans and robots represents the fundamental paradigm for the future coexistence of artificial intelligence (AI) agents and humans \cite{zhang2024emergence,zhang2019symmetrical}. This paradigm necessitates that robots demonstrate advanced cognitive abilities~\cite{clark1999towards} and intelligent behaviors~\cite{rahwan2019machine}, thereby enabling them to address intricate issues prevalent in human society. Cognitive robots~\cite{levesque2008cognitive}, which have significantly benefited from the swift advancements in AI, are capable of providing assistance by predicting human intentions through multi-modal perception \cite{sun2023neighbor}. This has emerged as a primary method for robots to serve human society. Consequently, the accurate comprehension of human complex intentions and the execution of tasks that align with human values~\cite{rokeach2008understanding} are pivotal research problems in the field of cognitive robotics.

\begin{figure}[tb]
  \vspace{3pt}
  \centering
  \includegraphics[width=\linewidth]{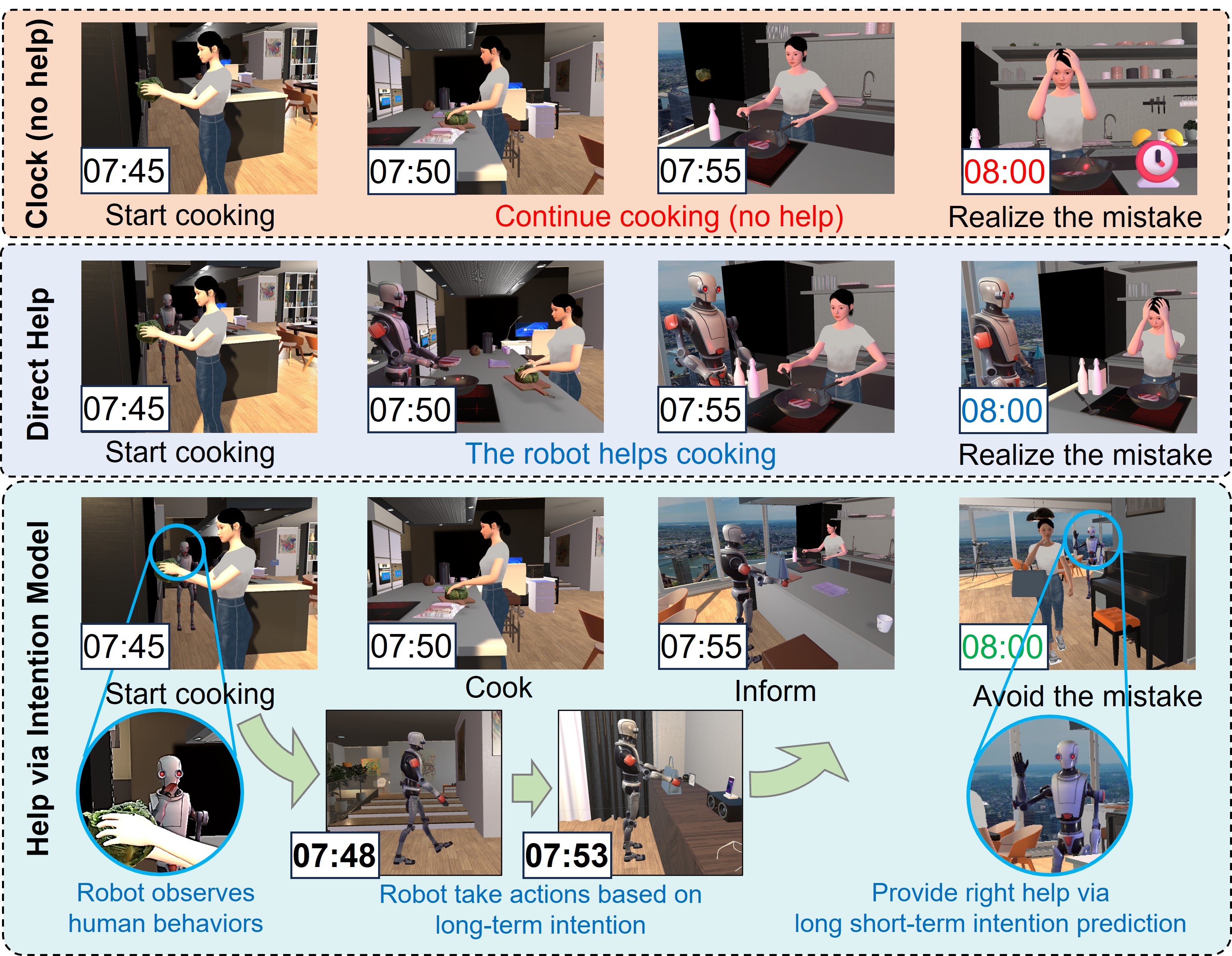}
  \vspace{-18pt}
  \caption{Comparison between intention model-based help and other forms. The top block denotes ``no help'' scenarios, in which the people could forget important issues. The middle block denotes ``direct help'' scenarios, in which the people may receive inappropriate help and fail to finish the more important thing. The bottom block denotes the ``help via intention model'' scenario, in which the robot can infer the long-term intention and warn the short-term mistake, thus providing appropriate service for people. }
  \label{fig:teaser}
\end{figure}

At present, a significant portion of research on intention understanding~\cite{blakemore2001perception,fogassi2005parietal,tomasello2023having} concentrates on immediate intentions by analyzing motion sequences. This approach aids in accomplishing tasks in which humans may need help. It operates on the assumption that a human’s immediate intention aligns with their values that determine the human behavioral patterns~\cite{schwartz2012overview}, without considering the potential for inappropriate actions or even mistakes. For example, an individual engrossed in complex cooking tasks might forget an appointment with a friend. In such a scenario, an intelligent robot is supposed to recognize that continuing the cooking tasks is not the optimal choice, and suggest that the human halt their current activity for attending the appointment. On the contrary, if the robot merely assists with the cooking tasks based on its observations without suggesting the appointment through more inference, it could be perceived as lacking intelligence. In practice, even the advanced large language models, such as ChatGPT~\cite{aljanabi2023chatgpt} and GPT-4~\cite{openai2023gpt4}, only account for the immediate situation in human-robot interaction, thus ignoring the influence of high-level human values. 
These AI models, as well as cognitive robots driven by these models, will often provide suboptimal assistance or even no assistance when the immediate actions of humans accidentally violate the human values.

To address the aforementioned issues, robots should be able to understanding complex human intentions based on human behavioral patterns. Specially, human intention is decomposed into the long-term intention which aligns with human values, and the short-term intention which represents the immediate action intention. Autonomous household robots are aiming at providing reasonable services when living with humans in a household environment, so they should be able to predict this kind of human intentions.
We refer to this kind of intention prediction mechanism as ``long short-term intention prediction''. 
This task introduces two challenges: firstly, building a long short-term intention model via observation to represent both human values and immediate action intentions; secondly, determining whether the action-based short-term intention conflicts with the value-based long-term intention. The first challenge necessitates a substantial amount of lifelog data for training the intention model. However, collecting this type of data is challenging, and there is no readily available dataset that supports this task. The second challenge involves predicting both the short-term intention and the long-term intention, and assessing whether these two types of intentions conflict with each other. In summary, the aforementioned intention model suitable for representing complex human intentions should be constructed, by which the robot can provide meaningful assistance services.

In this paper, we construct a continuous, long-term human life recording dataset by combining human data synthesis with human data recording. This dataset supports model training and testing, providing the fundamental data required to address the proposed intention prediction task. We propose to build a long short-term intention model for robots to understand human behaviors, which allows the robot's real-time detection of whether the anticipated outcomes of short-term human behavior align with the value-based long-term behavioral patterns. This further enables robots to provide reasonable suggestions as early as possible before the potential adverse consequences emerge, as shown in \autoref{fig:teaser}.

Our contributions are three-fold:
\begin{itemize}
    \item We construct a dataset for long short-term intention prediction tasks by combining large language model-based data synthesis with virtual simulation-based data recording. 
    \item We define the ``long short-term intention prediction'' task, and build a long short-term intention model to tackle this task, which allows the robot to learn long-term behavioral patterns of humans.
    \item Experimental results verify that the proposed intention model can predict complex intentions embedded in human behaviors, which might help indicate potential suboptimal behavior and even mistakes.
\end{itemize}

\section{RELATED WORKS}

\subsection{Intention Prediction}

The prediction of human intentions has long been a focal point in the field of cognitive robotics. Corresponding datasets involving human actions and intentions are needed to ensure the implementation of research. 

Actions are the basic units of human behaviors. Therefore, the study of human behavior often involves segmenting continuous motion sequences into discrete actions. This approach is exemplified by the AVA dataset~\cite{gu2018ava}. Concurrently, the interactions between humans and objects provide a wealth of information about human behavior. This has led to the creation of datasets like LEMMA~\cite{jia2020lemma}, which offers action labels for human-object interactions, thereby facilitating the understanding of multi-agent, multi-task activities in everyday life. As for robots, ego-centric tasks~\cite{jia_egotaskqa_2022} are crucial for understanding human behavior, for the reason that robots often perceive their environments from an ego-centric viewpoint. Building a long-term human life dataset regarding the robot's ego-centric viewpoint should be necessary for household robots.

Action intention prediction has been explored by some researchers.
Puig et al. introduced a simulator known as VirtualHome-Social, which incorporates a challenge related to intention prediction~\cite{puig2020watch}. The authors devised five tasks, each representing a different type of household activity. These tasks primarily concentrate on the spatial relationships between objects, without taking into account the specific fluent representations of the objects. So the intentions encoded in the long-term change of environmental states are not easy to be recognized.

More specifically, human actions can be bifurcated into categories such as intentional and unintentional actions. Epstein et al. posited that the velocity of an action is intrinsically linked to its intentionality~\cite{epstein_oops_2020}. Actions executed at a slower pace are likely to be intentional, whereas actions performed swiftly are likely to be unintentional~\cite{caruso_slowmotion_2016}. Accordingly, Epstein et al. introduced the OOPS dataset for the prediction of unintentional actions~\cite{epstein_oops_2020}. This dataset comprises 20,338 videos and is accompanied by a self-supervised algorithm that learns representations of intention. 

Even though the action intentions have been studied for a long time, the long-term intentions are merely mentioned by existing research works. An the long-term intention and short-term intention are often interconnected, which make it hard to predict true intentions of humans.
%Given the intricate intentions inherent in human actions, a robot necessitates a cognitive structure~\cite{kotseruba_review_2016} to understand human intentions.

\subsection{Intention-Aware Task Planning}

The capability of intention prediction will facilitate the downstream tasks like task planning.
Various methodologies have been proposed to articulate the planning problem, such as the planning domain definition language (PDDL)~\cite{aeronautiques1998pddl} and probabilistic PDDL (PPDDL)~\cite{younes2005first}, relational dynamic influence diagram language (RDDL)~\cite{sanner2010relational}, and behavior domain definition language (BDDL)~\cite{li2021igibson}.
In relatively complicated household environments, hierarchical planners can be employed. 
For instance, with the prediction of partners, an agent can plan how to assist its partner with partial observation~\cite{puig2020watch}. 

The planning process is often divided into high-level planning and low-level planning, which inspire us to study the proposed ``long short-term intention prediction'' problem with a hierarchical method. The widely-used long short-term memory \cite{hochreiter1997long} in deep learning field also support the effectiveness of long short-term mechanism, so we propose to develop a long short-term intention model to address the challenges in human intention prediction.

\section{LONG SHORT-TERM INTENTION PREDICTION}

\subsection{Problem Definition}

Value alignment is typically required in human-robot collaboration~\cite{yuan2022situ}, as it fosters trust between humans and robots. However, this becomes more complex in the context of household scenarios. However, even though robots can follow human decisions in every situation, the robots cannot be treated as fully intelligent. An obvious reason is that humans could be wisdom in a long run, but they might make mistakes in various immediate actions. An intelligent robot is expected to recognize what actions of humans should help and what actions of humans should be corrected from a long-term perspective.

Many efforts have been spent on such challenges to predict human attention, intention, and immediate collaboration~\cite{puig2020watch, fathi2013modeling, nan2020learning}.
But humans' immediate choices and actions may be suboptimal or even counterproductive when viewed in light of long-term behavioral patterns driven by human values, which has not been fully studied.

In the context of human-robot interaction in household environments, robots are expected to deal with this kind of problems, which is defined as \textbf{``long short-term intention prediction''}. In this task, robots need to predict both long-term and short term intentions of humans, and detect whether potential conflicts exist between different intentions.
For instance, \autoref{fig:teaser} shows an example where human's immediate action (cooking) is in conflict with her long-term behavioral patterns (going to work). An intelligent robot is supposed to align with the long-term intentions, rather than directly assisting humans based on short-term intentions, which may lead to  ineffective assistance or even destruction.

\subsection{Challenges}
The aforementioned ``long short-term intention prediction'' task involves the concurrent prediction of long-term and short-term intentions, as well as the identification of potential conflicts between these intentions.
To tackle this problem, one basic requirement is to parse human actions and environmental states (such as states of objects in the house).
Since object detection and action recognition are mature techniques, we will concentrate on the cognition part without the consideration of processing raw perception data (i.e., recognizing human action from raw images).
We attribute the rest of this problem to the following three challenges.

\subsubsection{Long-Term Observation}
We argue that humans' long-term behavior patterns are influenced by their values, while such patterns are different from person to person. The complex human intentions are encoded in the long-term behavior patterns of humans, so a dataset about human long-term life data is crucial in studying human intentions.

However, as mentioned in the related works, most datasets focus on short-term action data in several minutes or hours.
They care about learning universal patterns among a community rather than individual patterns.
As a result, they may have a lot of data from different people but only a few records per person.

Practically, to understand long-term behavior patterns, it is necessary to have long-term observations of humans individually, typically more than a fortnight.
Thus, we list long-term observation as the first challenge.
The observation should track each person as well as the surrounding environment for a relatively long time.
The expected observation data set $\mathcal{O}$ is denoted in \autoref{eq:observation}.
\begin{equation}
  \mathcal{O}_{p} = \left \{\mathcal{A},\mathcal{S},\mathcal{L}\right | t\in \left [ t_0, t_n \right ] , p \in \mathcal{P} \}   
  \label{eq:observation}
\end{equation}
where $\mathcal{A}$ denotes the action set observed from $t_0$ to $t_n$, $\mathcal{S}$ denotes the set of environment states observed from $t_0$ to $t_n$, $\mathcal{L}$ denotes the label set of each action-state data pair, $p$ denotes the index of the observed participant, and $\mathcal{P}$ denotes the participant set.
Note that the label set $\mathcal{L}$ could be the intention of each action annotated by humans or other useful information.

\subsubsection{Intention Modeling}
Numerous facets intertwine to define the complex concept of human intention, encompassing biological imperatives, psychological yearnings, ingrained behavioral patterns, and intricate personality traits, etc.
Some facets take a long time to be observed.
Modeling human intention in this task imposes a significant expectation upon the robot: the ability to predict human intentions over a long temporal horizon. 
The robot learns a function $\mathcal{F}_p:\mathcal{O}_p\to\mathcal{I}_p$ based on the observed behaviors, where $\mathcal{I}_p$ denotes the encoded intentions of the observations.
In the realm of decision-making, a well-established conundrum emerges, wherein the robot must navigate the terrain of short-term gains (e.g., greedy algorithm) versus long-term benefits (e.g., resisting the allure of high-calorie foods while adhering to a dietary regimen). 
This dilemma serves as a litmus test for the robot's capacity to transcend conventional paradigms of intention prediction.

\subsubsection{Conflict Recognition}
Inferring the intentions of others stands as a pivotal capability within the repertoire of human beings. 
It empowers individuals to proactively save others from making mistakes by dispensing valuable advice in advance. 
In this challenge, the robot is tasked with replicating and responding to this ability.
For elucidation, we represent the sequence of actions required to realize a particular intention as $\Vec{a} = \left ( \mathbf{a_1}, \mathbf{a_2}, ...,\mathbf{a_n} \right ) $.
On the one hand, the robot is required to predict the short-term intention before humans complete the actions in $\Vec{a}$.
On the other hand, the robot is required to predict a long-term intention which reflects human values.
The recognition of conflicts between different intentions could be denoted as $R_{conf}$ in \autoref{eq:conflict}:

\vspace{-9pt}
\begin{equation}
R_{conf}=\left\{
\begin{aligned}
1 & , & \{\mathcal{D}(I^{long}_{ t_k}, I^{short}_{t_k})> \delta| \vec{O_{t}},t\in\left [ t_0,t_k \right ] \}\\
0 & , & \{\mathcal{D}(I^{long}_{ t_k}, I^{short}_{t_k}) \le \delta| \vec{O_{t}},t\in\left [ t_0,t_k \right ] \}
\end{aligned}
\right.
\label{eq:conflict}
\end{equation}
where $I^{long}_{ t_k}$ denotes the long-term intention detected at $t_k$, $I^{short}_{ t_k}$ the short-term intention detected at $t_k$, function $\mathcal{D}(\cdot)$ the difference between $I^{long}_{ t_k}$ and $I^{short}_{t_k}$, $\delta$ the threshold for reporting conflicts, $\vec{O_t}$ the sequence of observation from $t_0$ to $t_k$.

\section{METHOD}

\subsection{Data Collection}
To tackle the first challenge, we create two kinds of behavior collectors.
On the one hand, behavior data directly collected from humans has advantages in terms of quality, but has disadvantages in the cost and collection difficulty.
On the other hand, large language models (LLMs) are used for data generation and data annotation, which is cheap but has more uncertainty than human annotations~\cite{touvron2023llama, wang2022self}.
We launch the two approaches for data collection and create two datasets: a simulated human behavior dataset and a collected human behavior dataset.

\subsubsection{Simulated Human Behaviors}

We prompt ChatGPT to generate simulated human behavior data. Within the prompt, we provide a detailed task description, constraints on actions, a list of interactable objects, descriptions of rooms within the house, as well as human and object states. The actions and initial states are specified to ensure the creation of a coherent and logical behavior plan. More specifically, we compel ChatGPT to adopt distinct personalities that encompass behavioral habits and preferences. We collected 19 descriptions from 19 participants through a questionnaire. These personalities serve as references for ChatGPT to generate behaviors. Additionally, we offer a few behavior examples that strictly adhere to a standardized format when generating behaviors. Furthermore, we define the initial states of all interactive objects. These states may be altered by future behaviors, including changes in object quantities and statuses. ChatGPT is tasked with simulating the role of a specific participant, generating a list of formatted descriptions of their daily behaviors within a specified date range.

Note that the list of objects, actions, and rooms given to ChatGPT is consistent with the one used in the collected human behavior dataset. They also share the same data format.

\subsubsection{Collected Human Behaviors}

In order to collect human data, we propose a data collection system built on Unity3D. This system features a virtual apartment comprising 6 rooms and a total of 206 interactable objects, including a living room, kitchen, bathroom, bedroom, study, and hall. Participants are asked to simulate their daily routines within this virtual apartment. They immerse themselves in the scene from a first-person perspective and can modify the states of the objects in the apartment through interactions. Participants are asked to select their actions using a user interface as if they were residing in their real-world homes. They specify the corresponding objects and label their action intentions (i.e., the short-term intentions).
After initializing objects with predetermined initial states, identical to those provided to ChatGPT, the system diligently records a comprehensive set of behavior data. This includes the action taken, the associated intention, the action's start time, its duration, the involved object, the resulting state of that object after the action, and the participant's position within the apartment. All collected data is stored in JSON format.
In anticipation of potential challenges such as insufficient memory or system downtime during the long-term data collection sessions, we have also implemented a breakpoint resume function. This feature allows the system to restore the states right up to the point of the last shutdown.

\begin{figure}[tb]
  \centering
  \vspace{5pt}
  \includegraphics[width=\linewidth]{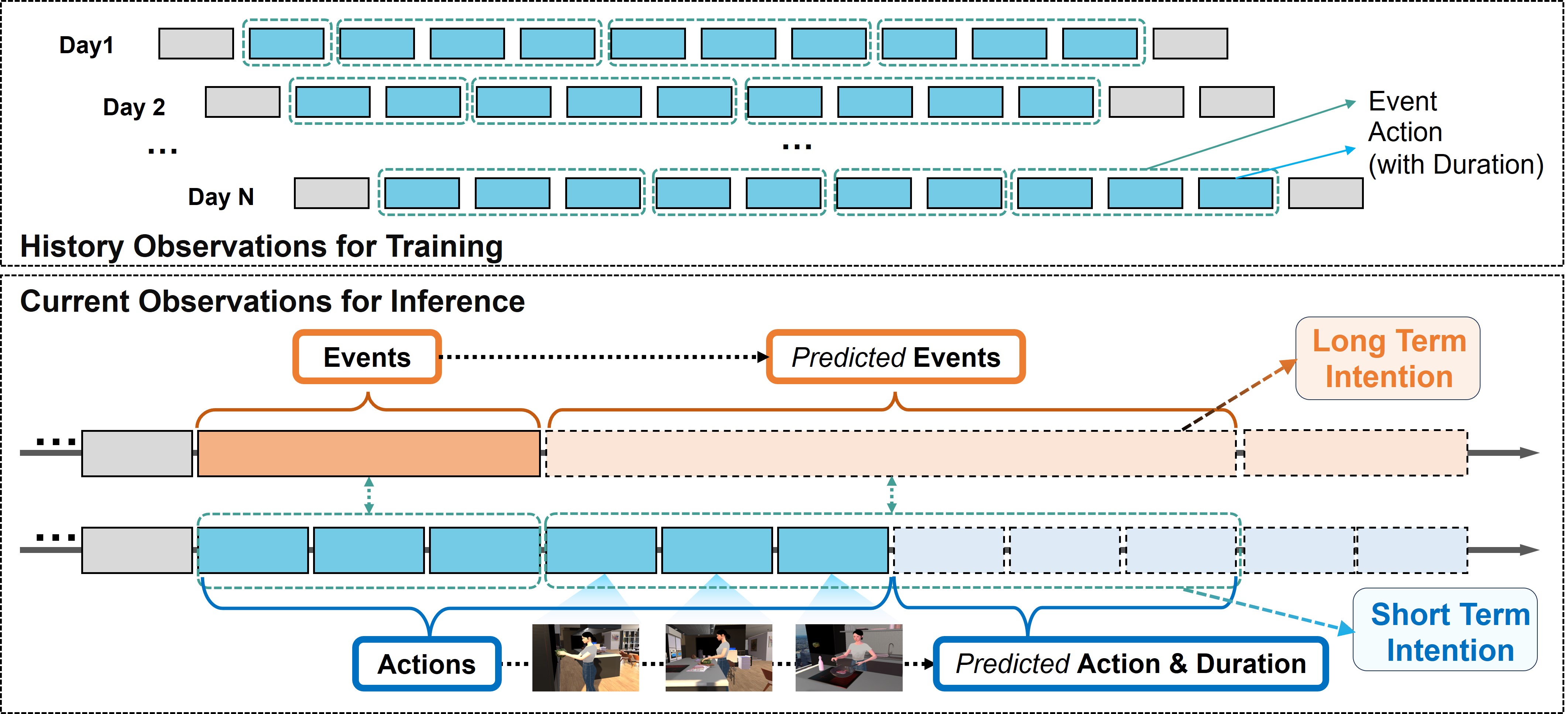}
  \vspace{-18pt}
  \caption{The long short-term intention model. The model involves the modeling for value-based long-term intention and the action-based short-term intention. This model allows the robot to decompose the complex human intention into long-term and short-term intentions, thus handling the potential human mistakes when short-term behaviors violate the long-term behavioral patterns.}
  \label{fig:system_pipeline}
  \vspace{-12pt}
\end{figure}

\begin{figure*}[tb]
  \centering
  \includegraphics[width=\linewidth]{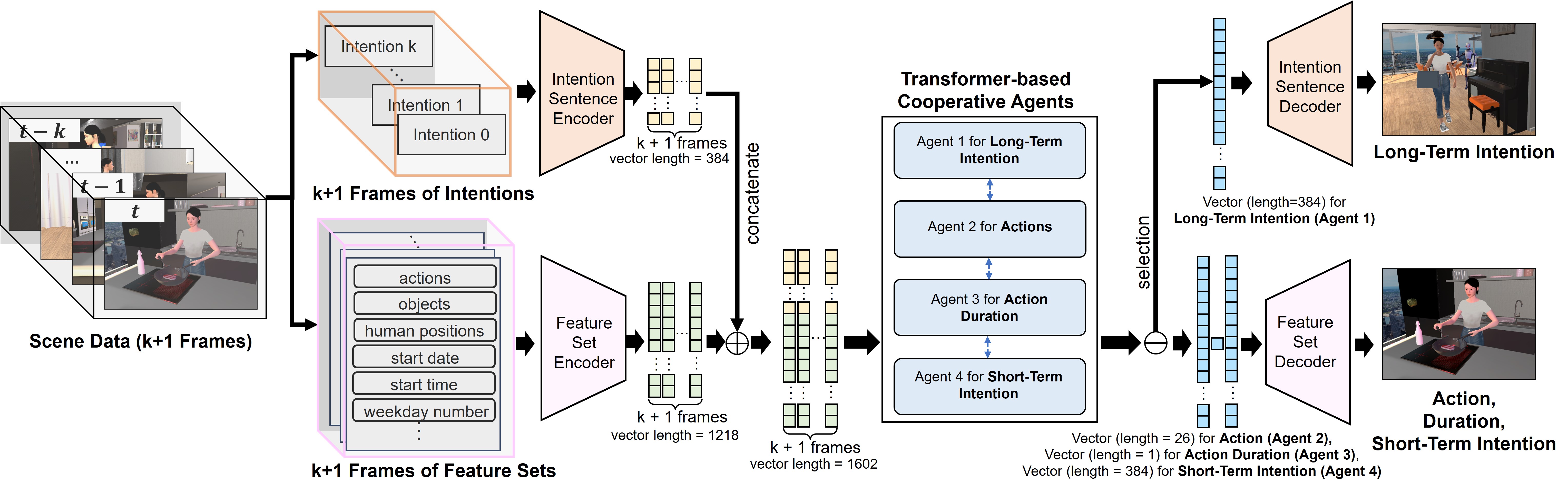}
  \vspace{-18pt}
  \caption{System pipeline. The temporal observation data is serialized before being fed into the neural networks. Intentions are singled out for feature encoding and decoding. The output of the system includes the prediction of actions, durations, short-term intentions, and long-term intentions.}   \vspace{-6pt}
  \label{fig:algorithm}
\end{figure*}

\subsection{Long Short-Term Intention Model}

Since intention prediction is a cognitive process, we refer to cognitive architectures when building the intention model. 
There has been a variety of cognitive architectures such as Belief-Desire-Intention (BDI)~\cite{harnad1990symbol}, Learning Intelligent Distribution Agent (LIDA)~\cite{faghihi2012lida},
Simulation of the Mental Apparatus \& Applications (SiMA)~\cite{schaat2015interdisciplinary}, Adaptive Control of Thought-Rational (ACT-R)~\cite{ritter2019act}, etc. 
AI models demonstrate high-level intelligence when incorporating values~\cite{peng2023tong}. Consequently, a value-based intention model holds significant potential for managing complex tasks. Meanwhile, human values are often reflected in long-term behavior patterns, which manifest the long-term intention prediction in human-robot interaction.
Following these, we propose our long short-term intention model.

Our modeling methodology addresses the task of long-short term intention through  the combination of action-based short-term and value-based long-term behavioral analyses, as shown in \autoref{fig:system_pipeline}.
Specially, this model takes human behavior information and environmental states into account, including but not limited to the action name, its start time and start date, its duration, the related objects, state changes caused by the action, and human states.
Observations of human behaviors are indexed by their start time.
Using a feature-extract function $\mathcal{F}_{extract}(\cdot)$, the observations with multi-source heterogeneous data are converted into a high-dimensional feature domain.
Each observation corresponds to a vector in the feature domain, denoted as $\Vec{obs}$ in \autoref{eq:feature_vector}.
\begin{equation}
  \Vec{obs_t} = \mathcal{F}_{extract}(O_t),t\in\left [ t_0,t_k \right ]
  \label{eq:feature_vector}
\end{equation}
The intention $\mathcal{I}_{t_k}$ starting from $t_k$ may relate to a continuous sequence of observations denoted as $\Vec{O}|_{t_k}$.
This model puts special attention on the time feature and consists of three layers of attention:

\begin{itemize}
    \item \textbf{L1: action.} It models the transitions between $\Vec{obs}$ to predict the next $\mathbf{a}_{t_{i+1}}$ along with the possible $\Vec{obs}_{t_{i+1}}$ and the duration of $\mathbf{a}_{t_{i+1}}$. This layer serves as a foundation for L2.
    \item \textbf{L2: short-term intention.} The prediction of future actions along with the observed history actions formulates an action list $\Vec{a} = \left ( \mathbf{a_1}, \mathbf{a_2}, ...,\mathbf{a_n} \right ) $. This layer models the short-term intention $I^{short}$ behind $\Vec{a}$, which is also known as immediate action intention.
    \item \textbf{L3: long-term intention.} It predicts a list of possible intentions $\Vec{I}^{long}$  using observation over an extended time window, considering daily routines and past intentions.
\end{itemize}

This model thereby facilitates the detection of conflicts between short-term and long-term intentions. 
With the aid of this proposed intention model, robots can gain a profound understanding of human behaviors, representing a pivotal advancement in the pursuit of holistic human-AI interaction.

\subsection{System Implementation}

The overall structure of the system implementation is shown in \autoref{fig:algorithm}.

\textbf{Data pre-process.}
The observation data is pre-processed before being fed into the neural networks.
In the pre-processing module, observation data $O_t$ is serialized into a multidimensional vector. 
As the LLM may generate ghost behavior data, we double-check the simulated data before serialization.
Most illegal data is fixed in an automated manner and the rest is discarded.
Actions, objects, and human positions are encoded into binary codes during the serialization.
Start time, start date, and weekday number are emphasized as separate features.
Action durations and object states are encoded as vectors and are concatenated with the aforementioned features.
Intentions of the observation are encoded into a feature domain using Sentence-BERT~\cite{reimers2019sentenceBert}.

\textbf{Model structure.}
To handle the complex intentions (including L1 to L3 intentions) of human behaviors, we propose to use 4 agents to make up of the whole intention model, and every agent corresponds for one prediction element (action, action duration, short-term intention and long-term intention) cooperate with other agents to produce the final prediction result. Within this model, every agent adopts a transformer-based structure, which is just a shallow transformer with only one encoder layer and one decoder layer. 
%The basic workflow is shown in \autoref{fig:algorithm}.

We also propose a conflict detection mechanism. We compare the similarity of the predicted long-term intention and the short-term intention. Note that we predict a list of potential long-term intentions in consideration of human uncertainties. If the short-term intention is not similar to any of the long-term intentions in the predicted list, the agent sends a query and proposes to help.

\section{EXPERIMENTS}

\subsection{Experimental Setup}
Four transformer-based agents are trained to learn actions, action durations, short-term intentions, and long-term intentions.
For each agent, we first train it using the simulated human behavior dataset (based on 19 different personalities) generated by GPT-3.5 to learn the general behavioral patterns among people.
This refers to the process of learning commonsense knowledge of people.
Afterwards, the agent gets fine-tuned with data sequences of every participant separately, so the agent can learn the different patterns of different participants. 
It refers to the process of personalization.
We finally test the performance of each agent on the test set of the corresponding participant, and compare the performance with the baseline models.

We use 70\% of the simulated dataset to train the agents, while 10\% for validation and 20\% for testing.
The test set contains data from 8 participants.
Four of them are simulated using GPT-4 (P01-P04) while the other four are human participants (P05-P08).
For each participant, we fine-tune the basic model using 70\% of his/her data and 10\% for validation.
The remaining 20\% data is used for testing.

\begin{table*}[t]
%\vspace{-20pt}
\caption{Model performance on data from different participants. For duration, the lower the better; for action, short-term intention, and long-term intention, the higher the better. Note that the ``End to end'' method cannot distinguish the short-term and long-term intentions, so we leave it empty for long-term intention prediction.  }
\vspace{-21pt}
\label{tab:baseline_compare}
\begin{center}
\setstretch{0.95}
\resizebox{\linewidth}{!}{
\begin{tabular}{cc|c|ccc|ccc|ccc}
\hline
\multicolumn{2}{c|}{\multirow{2}{*}{Configuration}} & duration ($\downarrow$)  & \multicolumn{3}{c|}{action ($\uparrow$)} & \multicolumn{3}{c|}{short-term intention ($\uparrow$)} & \multicolumn{3}{c}{long-term intention ($\uparrow$)} \\ \cline{3-12} 
\multicolumn{2}{c|}{}     & relative error (\%) & top-1 (\%)   & top-3 (\%)   & top-5 (\%)   & top-1 (\%)      & top-3 (\%)     & top-5 (\%)     & top-1 (\%)     & top-3 (\%)     & top-5 (\%)    \\ \hline
\multirow{8}{*}{End to end}     
                          & P01    & 278.67   & 14    & 56    & 81    & 9       & 30      & 40      & -      & -      & -     \\
                          & P02    & 303.83   & 16    & 37    & 45    & 13       & 21      & 27      & -      & -      & -     \\
                          & P03    & 198.40   & 17    & 38    & 60    & 4       & 11      & 12      & -       & -     & -     \\
                          & P04    & 121.73   & 19    & 43    & 63    & 1       & 11      & 25      & -      & -     & -     \\
                          & P05    & 81.73   & 18    & 32    & 42    & 4       & 21      & 23      & -      & -      & -     \\
                          & P06    & 84.17   & 23    & 52    & 63    & 3       & 15      & 26      & -      & -       & -      \\
                          & P07    & 43.15   & 8    & 32    & 36    & 56       & 64      & 64      & -      & -      & -     \\
                          & P08    & 101.25   & 13    & 35    & 45    & 13       & 44      & 57      & -      & -      & -     \\ 
                          \hline
\multirow{8}{*}{Ours}     & P01    & 68.35   & 81    & 95    & 97    & 62       & 72      & 83      & 73      & 79      & 79     \\
                          & P02    & 80.39   & 65    & 73    & 75    & 63       & 75      & 77      & 62      & 77      & 79     \\
                          & P03    & 65.07   & 57    & 67    & 72    & 41       & 57      & 61      & 30       & 48      & 52     \\
                          & P04    & 56.10   & 70    & 84    & 85    & 66       & 71      & 79      & 64      & 71      & 74     \\
                          & P05    & 10.82   & 59    & 80    & 82    & 76       & 86      & 88      & 76      & 83      & 90     \\
                          & P06    & 15.52   & 42    & 63    & 65    & 27       & 55      & 60      & 32      & 50       & 50      \\
                          & P07    & 40.24   & 44    & 60    & 64    & 58       & 79      & 89      & 84      & 96      & 96     \\
                          & P08    & 14.11   & 47    & 65    & 68    & 39       & 82      & 87      & 82      & 89      & 89     \\ 
                          \hline
\end{tabular}
}
\end{center}
\vspace{-18pt}
\end{table*}

\subsection{Model Training}
\subsubsection{Data}
For each of the aforementioned agents, we train the model respectively.
(i) The same input data formulation is utilized for predicting actions and durations: the action, its start time, its duration, the involved objects, the states of the objects after the action, and the participant's position within the apartment.
The ground truth is the last observation of the input sequence.
(ii) The input data for predicting short-term intentions is similar to that used to predict actions. The ground truth has one more label: the intentions related to the actions.
(iii) The input data for predicting long-term intentions includes the labeled intentions.

\subsubsection{Loss}
We utilized cross-entropy loss when training the action model.
For action durations, mean square error is utilized as loss.
For the short and long-term intentions, a cosine embedding loss is utilized to compare the similarity between the predicted intention and the ground truth.

\subsection{Model Test}
\subsubsection{Test Set}
We proposed a test set with two kinds of data: simulated human behaviors generated by GPT-4 (P01-P04) and collected human behaviors from real humans (P05-P08).

\subsubsection{Baseline}
The baseline model has a similar basic structure but is trained end-to-end, and thus, it does not follow the intention model we proposed.
As long short-term intention prediction is a new task, we choose this baseline to evaluate the effectiveness of our proposed long short-term intention model.

\subsubsection{Metric}
The relative error between the predicted duration and the ground truth is utilized to evaluate the duration prediction: $Error(pred, gt) = |(pred-gt)/(gt+\epsilon)| $, where $pred$ represents the predicted duration and $gt$ represents the ground truth. $\epsilon$ is a small compensation value in case the ground truth is zero.
Actions are seen as discrete symbols from a closed set $\mathcal{A}$ and are evaluated using top-1, top-3, and top-5 accuracies.
The ground truth of intentions is described in natural language.
To make the result clearer, we build an intention set $\mathcal{I}_{P_i}$ for each participant that contains all existing intentions of $P_i$, and compare the similarity of the predicted intention with intentions in $\mathcal{I}_{P_i}$. If the most similar one is the ground truth, this prediction is counted for top-1 accuracy. The top-3 and top-5 accuracies are calculated similarly.

\subsection{Results.}

We compare the performance of the end-to-end model and our intention model among the two kinds of test data.
The result is shown in \autoref{tab:baseline_compare}.
Overall, our method based on the long short-term intention model has better performance than the end-to-end method.
The table also shows that the learning difficulty is different from person to person, because different participants have their unique behavior patterns.
For example, P03 and P06 are considered difficult cases according to the result.

\textbf{Duration.}
The results of the collected data are better than that of the GPT-4 data, which indicates that the collected data has a smoother distribution than simulated data.

\textbf{Action.}
There are 26 kinds of actions in the dataset.
The accuracies of simulated data are higher than that of the collected data.
This indicates that the simulated data has better consistency while the collected data from real humans has more randomness.

\textbf{Short-term intention} and \textbf{long-term intention.}
The performance of predicting long-term intention is slightly higher than short-term intention among a majority of participants.
For most participants, the top-5 accuracy is around or over 80\%.
Therefore, we include 5 intentions in the long-term intention list.
If the robot finds the user's short-term intention is not in the list, it raises a query to remind the user if anything has been missed.

\subsection{Discussion}
\textbf{System application.}
With the prediction of action sequence, the robot further predicts the short-term intention and a list of long-term intentions of the user.
Then it compares the similarity of short-term and long-term intentions, considering the duration of actions.
This should be of importance for autonomous household robots to understand humans and provide appropriate assistance.

\textbf{Comparison with GPT.}
We launched a qualitative comparison between our method and LLMs (i.e., ChatGPT).
We found that LLM is good at inferring short-term intentions but does not care much about long-term intentions. Therefore, the combination of LLMs and specific designs such as long-term intention mechanism might be a potential approach to reach a more intelligent level.

\textbf{Limitations.}
Since the focus of our work is concentrated on the intention prediction model, and the input and output data are both symbolic, so the assumption is the robot can perceive the world and take actions accurately. However, in real scenarios, the noisy environment cannot be omitted. Next step, the research is planning to design and implement experiments in complex physical household environments.
   
\section{CONCLUSION}

Our work underscores the importance of imbuing robots with intelligent behavior, particularly in their ability to understand the complex intentions of humans. 
In response to this challenge, we introduced a novel task called ``long short-term intention prediction'', which involves proactively identifying humans' short-term intentions and long-term intentions, as well as the recognition of potential conflicts between long-term and short-term intentions. To find a solution, we constructed a dataset for training intention models, and gave a complete pipeline to foster robots in understanding human intentions. Experimental results have validated the effectiveness of the proposed long short-term intention model.

%%%%%%%%%%%%%%%%%%%%%%%%%%%%%%%%%%%%%%%%%%%%%%%%%%%%%%%%%%%%%%%%%%%%%%%%%%%%%%%%

% \bibliographystyle{IEEEtran}
% \bibliography{IEEEFULL}

\end{document}